\pdfoutput=1
\documentclass[11pt]{article}

\usepackage[final]{acl}

\usepackage{times}
\usepackage{latexsym}

\usepackage[T1]{fontenc}
\usepackage[utf8]{inputenc}

\usepackage{microtype}

\usepackage{inconsolata}

\usepackage{booktabs}
\usepackage{adjustbox}    % or use graphicx for resizebox
\usepackage{caption}
\usepackage{supertabular}
\usepackage[most]{tcolorbox}

\usepackage{hyperref}
\usepackage{url}
\usepackage{diagbox}
\usepackage{algorithm}
\usepackage{algpseudocode}
\usepackage{amsmath}
\usepackage{amssymb}
\usepackage{amsmath}
\usepackage{arydshln}
\usepackage{ascii}
\usepackage{bm}
\usepackage{booktabs}
\usepackage{colortbl}
\usepackage{color}
\usepackage{CJK}
\usepackage{dsfont}
\usepackage{enumitem}
\usepackage{forest}
\usepackage{graphics}
\usepackage{graphicx}
\usepackage{latexsym}
\usepackage{multirow}
\usepackage{mwe}
\usepackage{makecell}
\usepackage{microtype}
\usepackage{subfig}
\usepackage{times}
\usepackage{tikz}
\usepackage{tabularx}

\usepackage{url}
\usepackage{wrapfig}
\usepackage{tcolorbox}
\usepackage{xcolor}
\definecolor{comments}{RGB}{0, 150, 0}

\definecolor{highlight}{RGB}{200, 0, 0}

\title{On LLM-Based Scientific Inductive Reasoning Beyond Equations}

\title{On LLM-Based Scientific Inductive Reasoning Beyond Equations}

\author{
\textbf{Brian S. Lin}$^{1}$\thanks{Equal contribution.}
\textbf{Jiaxin Yuan}$^{2}$\footnotemark[1]
\textbf{Zihan Zhou}$^{3}$\footnotemark[1]
\textbf{Shouli Wang}$^{4}$\footnotemark[1]
\textbf{Shuo Wang}$^{1}$\thanks{Corresponding authors.}\\
\textbf{Cunliang Kong}$^{1}$
\textbf{Qi Shi}$^{1}$
\textbf{Yuxuan Li}$^{1}$
\textbf{Liner Yang}$^{2}$\footnotemark[2]
\textbf{Zhiyuan Liu}$^{1}$\footnotemark[2]
\textbf{Maosong Sun}$^{1}$ \\
$^{1}$Dept. of Comp. Sci. \& Tech., Institute for AI, BNRist Center, Tsinghua University \\
Jiangsu Collaborative Innovation Center for Language Ability, Jiangsu Normal University \\
$^{2}$Beijing Language and Culture University \\
$^{3}$Xiamen University \quad 
$^{4}$Harbin Institute of Technology \\
\texttt{caish25@mails.tsinghua.edu.cn} 
}

\begin{document}
\maketitle
\begin{abstract}
As large language models (LLMs) increasingly exhibit human-like capabilities, a fundamental question emerges: How can we enable LLMs to learn the underlying patterns from limited examples in entirely novel environments and apply them effectively? This question is central to the ability of LLMs in inductive reasoning. Existing research on LLM-based inductive reasoning can be broadly categorized based on whether the underlying rules are expressible via explicit mathematical equations. However, many recent studies in the beyond-equations category have emphasized rule design without grounding them in specific scenarios. Inspired by the parallels between inductive reasoning and human scientific discovery, we propose the task of LLM-Based Scientific Inductive Reasoning Beyond Equations and introduce a new benchmark, SIRBench-V1, to evaluate the inductive reasoning abilities of LLMs in scientific settings. Our experimental results show that current LLMs still struggle with this task, underscoring its difficulty and the need for further advancement in this area.\footnote{The open-source code and data can be found at \url{https://github.com/thunlp/SIR-Bench}.}
\end{abstract}

\section{Introduction}

\begin{table*}[ht]
  \centering
  \resizebox{\textwidth}{!}{ % 表格宽度匹配双栏总宽
  \begin{tabular}{
    >{\centering\arraybackslash}m{3.5cm} % Benchmark
    >{\centering\arraybackslash}m{3.4cm} % Task Type
    >{\centering\arraybackslash}m{2.7cm} % Related to Scientific Discovery
    >{\centering\arraybackslash}m{3.1cm} % Beyond Equations
    >{\centering\arraybackslash}m{2.9cm} % Rule-identifiable
    >{\centering\arraybackslash}m{1.8cm} % # Instances
    >{\centering\arraybackslash}m{2.5cm} % Sequence Length
  }
    \toprule
    \textbf{Benchmark} & \textbf{Task Type} & \textbf{Related to Scientific Discovery} & \textbf{Beyond Mathematical Equations} & \textbf{Closed-Ended Questions} & \textbf{\#Instances} & \textbf{Sequence Length} \\
    \midrule
    MATDESIGN & HI & \checkmark & \checkmark & × & 50 & 250-1,000 \\
    TOMATO-Chem & HI & \checkmark & \checkmark & × & 51 & 100-600 \\
    ResearchBench & HI & \checkmark & \checkmark & × & 1,386 & unknown \\
    chaotic systems & SR & \checkmark & × & \checkmark & 131 & \textasciitilde100 \\
    SRSD & SR & \checkmark & × & \checkmark & 240 & 100-300 \\
    LLM-SRBench & SR & \checkmark & × & \checkmark & 239 & \textasciitilde100 \\
    MIRAGE & IR & × & \checkmark & \checkmark & 2,000 & 20-100 \\
    MIR-Bench & IR & × & \checkmark & \checkmark & 6,930 & 50-250 \\
    IOLBench & IR & × & \checkmark & \checkmark & 1,500 & 200-2,000 \\
    \midrule
    \textbf{SIRBench-V1 (Ours)} & IR & \checkmark & \checkmark & \checkmark & 710 & 500-3,000 \\
    \bottomrule
  \end{tabular}
  }
  \caption{Analysis of existing related benchmarks. \textbf{HI}: \textit{Hypothetical Induction}, \textbf{SR}: \textit{Symbolic Regression}, \textbf{IR}: \textit{Inductive Reasoning}. \textbf{Related to Scientific Discovery}: targets scientific problem-solving. \textbf{Beyond Mathematical Equations}: focuses on reasoning not reducible to equation fitting. \textbf{Closed-Ended Questions}: has deterministic answers for automatic evaluation. \textbf{\#Instances}: number of test examples. \textbf{Sequence Length}: input sequence length---crucial as scientific inductive reasoning often requires extracting information from extensive resources.
}
  \label{tab:benchmark-overview}
\end{table*}

In recent years, 
% the reasoning capabilities of large language models (LLMs) have shown significant improvements~\cite{Plaat2024ReasoningWL,Bubeck2023SparksOA}. 
many advanced reasoning models, including OpenAI o1~\cite{openai2024openaio1card} and DeepSeek-R1~\cite{DeepSeekAI2025DeepSeekR1IR}, have demonstrated strong {\em deductive reasoning} capabilities, especially as evidenced by their performance in mathematics and programming tasks. These tasks are typically characterized by concise problem descriptions, where the model is required to generate a long chain of thought~\cite{Wei2022ChainOT} to solve complex problems.

% \begin{figure*}[!t]
% 	\centering
% 	\includegraphics[width=0.5\linewidth]{imgs/task_example.pdf}
%     \caption{Our xxx}

% 	\label{fig:SIR}
% \end{figure*}

\begin{figure}[htbp]
  \centering
  \includegraphics[width=\columnwidth]{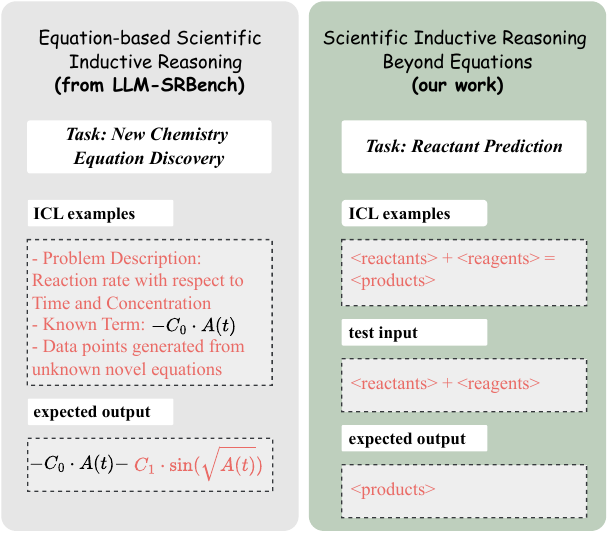}
  \caption{Illustrative comparison of scientific inductive reasoning: on the left, tasks focused on equation discovery~\cite{Shojaee2025LLMSRBenchAN}, and on the right, tasks representing broader forms of scientific induction beyond equation generation.}
  \label{fig:single_column}
\end{figure}

In contrast, {\em inductive reasoning}~\cite{hayes2010inductive} poses a different challenge, requiring models to infer general rules or structures from multiple specific observations~\cite{chollet2019measureintelligence,Yang2022LanguageMA}. Inductive reasoning involves making predictions about new scenarios based on existing knowledge or observed data~\cite{hayes2010inductive}. Inductive reasoning has been progressively recognized as a critical component for human-like cognitive modeling and the development of general artificial intelligence~\cite{Li2024MIRAGEEA}. However, current LLMs still exhibit notable shortcomings in inductive reasoning tasks~\cite{Li2024MIRAGEEA,Hua2025InductionBenchLF,Yan2025MIRBenchBL}. Even state-of-the-art models often fail to correctly infer abstract rules from observations and typically rely on memorizing rather than truly understanding the underlying concepts. 

Currently, artificial intelligence is increasingly regarded as a transformative paradigm in scientific discovery, with growing applications across disciplines such as physics, materials science, and chemistry~\cite{Xu2021ArtificialIA}. Against this backdrop, increasing attention has been paid to the inductive reasoning abilities of LLMs in scientific contexts recently~\cite{Yang2024MOOSEChemLL,Liu2025ResearchBenchBL,fang2025ainewtonconceptdrivenphysicallaw}. However, systematically leveraging reasoning models to enhance inductive tasks for scientific discovery remains largely underexplored.

While some scientific rules, such as the velocity formula of free fall, can be expressed mathematically, others, such as molecular structure-function relationships, are not readily amenable to such formulation. Under this criterion, we observe that existing LLM-based inductive reasoning research can be broadly categorized based on {\em whether the underlying rules can be formulated mathematically}. The first category comprises tasks that are mathematical equation-based, which are closely related to symbolic regression~\cite{Matsubara2022RethinkingSR,Gilpin2021ChaosAA}. Recent work has shown that LLMs can serve as equation generators or guide the equation discovery process ~\cite{Wang2024ExploringTM,Du2024LargeLM,Shojaee2024LLMSRSE,Shojaee2025LLMSRBenchAN,fang2025ainewtonconceptdrivenphysicallaw}. However, these tasks typically only cover cases where the underlying rules can be explicitly formulated as equations.
% A separate line of work targets tasks beyond mathematical equations. These works propose new inductive tasks and datasets from various perspectives~\cite{Hua2025InductionBenchLF,Tang2024MarsSI,Banatt2024WILTAM,Goyal2025IOLBENCHBL}, but many of them place emphasis on the construction of novel synthetic or low-frequency systems, which are not closely related to discovering scientific patterns in real-world scenarios. \comments{To Modify} 
A separate line of work targets tasks beyond mathematical equations, proposing new inductive tasks and datasets from various perspectives~\cite{Hua2025InductionBenchLF,Tang2024MarsSI,Banatt2024WILTAM,Goyal2025IOLBENCHBL}. However, many of these studies emphasize the creation of novel synthetic or low-frequency symbolic systems, which often have a limited connection to discovering scientific patterns in real-world scenarios.
Recent efforts under the AI4Science agenda are exploring more scientifically grounded settings where models emulate researchers by deriving insights or hypotheses from scientific materials~\cite{Yang2023LargeLM,Yang2024MOOSEChemLL,Liu2025ResearchBenchBL}. However, the reasoning processes of these studies often remain coarse-grained or open-ended, making robust automatic evaluation challenging.

To address these gaps, we propose to examine the capabilities of LLMs in {\em Scientific Inductive Reasoning Tasks Beyond Mathematical Equations}. To the best of our knowledge, high-quality and easy-to-evaluate datasets to directly investigate this problem are currently lacking. We have therefore created {\em SIRBench-V1}, a new benchmark consisting of a series of subtasks in chemistry and biology. In these subtasks, the underlying rules cannot be expressed through mathematical equations, yet they yield relatively deterministic answers.
We transform basic scientific resources from prior studies~\citep{grevsova2023genomic,Liu2024ExploringGL,Guo2023WhatCL,edwards-etal-2022-translation,Irwin2021ChemformerAP,Westerlund2024DoCD,westerlund2024constrained,Kim2018PubChem2U} into inductive reasoning tasks. Furthermore, to eliminate LLM memorization, we design counterfactual tasks that establish synthetic scientific rules for the models to reason with, rather than recall.

We follow several commonly adopted reasoning strategies for LLMs on the SIRBench-V1, including implicit and explicit reasoning, self-consistency~\cite{wang2022self}, and hypothesis refinement~\cite{Qiu2023PhenomenalYP}. By investigating the performance of several LLMs augmented with different reasoning strategies, we find that equation-free scientific inductive reasoning is highly challenging for modern LLMs. Gemini-2.5-Flash, the best-performing model, achieves an average accuracy of $43.81\%$ in our benchmark, while Claude-3.5-Haiku and GPT-4.1 demonstrate a lower average accuracy of $31.53\%$  and $32.41\%$, respectively. We also observe that using sophisticated reasoning strategies provides minimal performance improvement and, in some cases, even leads to performance decline. Using hypothesis refinement, Gemini-2.5-Flash, Claude-3.5-Haiku, and GPT-4.1 attain an average accuracy of $39.06\%$, $31.63\%$, and $33.25\%$, respectively. We believe this work will pave the way for a new and fruitful avenue of research in scientific discovery.

\paragraph{Contributions} In summary, the main contributions of this work are as follows:
\begin{itemize}
    \item We present SIRBench-V1, a new scientific inductive reasoning benchmark featuring authentic and counterfactual test examples from tasks in both biology and chemistry.
    \item We conduct evaluations using several representative LLMs in conjunction with diverse advanced inference strategies, the results of which demonstrate the capability boundaries of the examined LLMs.
	% \item
 %    We have provided a complete pipeline for the task, including the construction of the SIRBench-V1, the application of several reasoning methods, and ready-to-run evaluation scripts, etc.
	% We introduce the SIRBench-V1 benchmark, which includes representative tasks from chemistry and biology, and is designed to \comments{xxx}. 
    \item We derive several constructive findings for scientific inductive reasoning, such as a comparison between many-short-shot and long-few-shot learning approaches and an analysis of memorization, which we anticipate will be helpful for subsequent studies. 
\end{itemize}

\section{Related Work}
\subsection{Inductive Reasoning}
% As inductive reasoning gains attention in cognitive modeling and general AI, more studies are exploring how to systematically evaluate language models on such tasks. To this end, various benchmarks have been proposed to assess their performance in different inductive reasoning scenarios.
\paragraph{Benchmark}
Various benchmarks have recently been introduced to systematically evaluate these capabilities from multiple perspectives. \citet{Hua2025InductionBenchLF} evaluate the model's ability to infer string transformation rules from limited input-output examples. Bongard-OpenWorld~\cite{Wu2023BongardOpenWorldFR} examines conceptual induction and image classification in few-shot scenarios. \citet{Tang2024MarsSI} propose an embodied interactive environment requiring models to induce task rules and objectives. MIR-Bench~\cite{Yan2025MIRBenchBL} provides a many-shot in-context benchmark covering various function-based input-output pairs. WILT~\cite{Banatt2024WILTAM}, inspired by the Wason 2-4-6 task, evaluates multi-turn inductive reasoning and generalization capabilities. Additionally, benchmarks such as LINGOLY~\cite{Bean2024LINGOLYAB}, Linguini~\cite{Snchez2024LinguiniAB} and IOLBench~\cite{Goyal2025IOLBENCHBL}, derived from the International Linguistics Olympiad, challenge model generalization under low-resource language scenarios. 

\paragraph{Methods}
Beyond benchmark development, recent efforts have also explored structured frameworks to enhance inductive reasoning in LLMs, addressing limitations observed with chain-of-thought prompting and few-shot methods~\cite{Bowen2024ACE,Gendron2023LargeLM}. For instance, Chain-of-Language-Models~\cite{Yang2022LanguageMA} employs a modular pipeline integrating rule generation and verification. \citet{Qiu2023PhenomenalYP} combines LLMs with symbolic executors in a propose-verify-refine loop, significantly enhancing robustness. Similarly, the De-In-Ductive (DID)~\cite{Cai2024TheRO} simulates a human-like inductive-then-deductive reasoning sequence within a single prompt, enabling flexible strategy switching and improved cross-task generalization.
\subsection{Scientific Inductive Reasoning in LLMs}
\paragraph{Symbolic Regression}
Symbolic regression is a core approach for scientific discovery~\cite{Matsubara2022RethinkingSR,Gilpin2021ChaosAA}. It is valued for its ability to extract analytical expressions directly from data ~\cite{Angelis2023ArtificialII}. Recent studies have extended this paradigm by incorporating LLMs into the tasks. In materials science,  \citet{Wang2024ExploringTM} highlight its role in revealing underlying physical and chemical principles. \citet{Du2024LargeLM} propose a prompt-based framework using LLMs to generate candidate equations, offering greater flexibility than traditional methods. \citet{Shojaee2024LLMSRSE} treat equations as programs, guided by scientific priors. To support systematic evaluation, they then introduce LLM-SRBench, a multi-domain benchmark designed to evaluate LLMs’ true discovery capabilities.
 
\paragraph{Hypothetical Induction}
Hypothetical Induction has been recognized as a subtask of inductive reasoning~\cite{Norton2003ALS}, with growing interest in using LLMs to generate novel, valuable scientific hypotheses from background knowledge or observations. \citet{Kumbhar2025HypothesisGF} introduced a goal-driven dataset and evaluation framework in materials science, while \citet{Yang2023LargeLM,Yang2024MOOSEChemLL} constructed datasets for hypothesis generation in chemistry and social science. Researchbench~\cite{Liu2025ResearchBenchBL} further provides the first benchmark covering inspiration retrieval, hypothesis formulation, and ranking.

\section{SIRBench-V1: Task and Construction}
We curate 7 tasks, with 100 samples for each biology task, including synthetic tasks, and 30 samples for each chemistry task.
\begin{figure*}[!t]
	\centering
	\includegraphics[width=\linewidth]{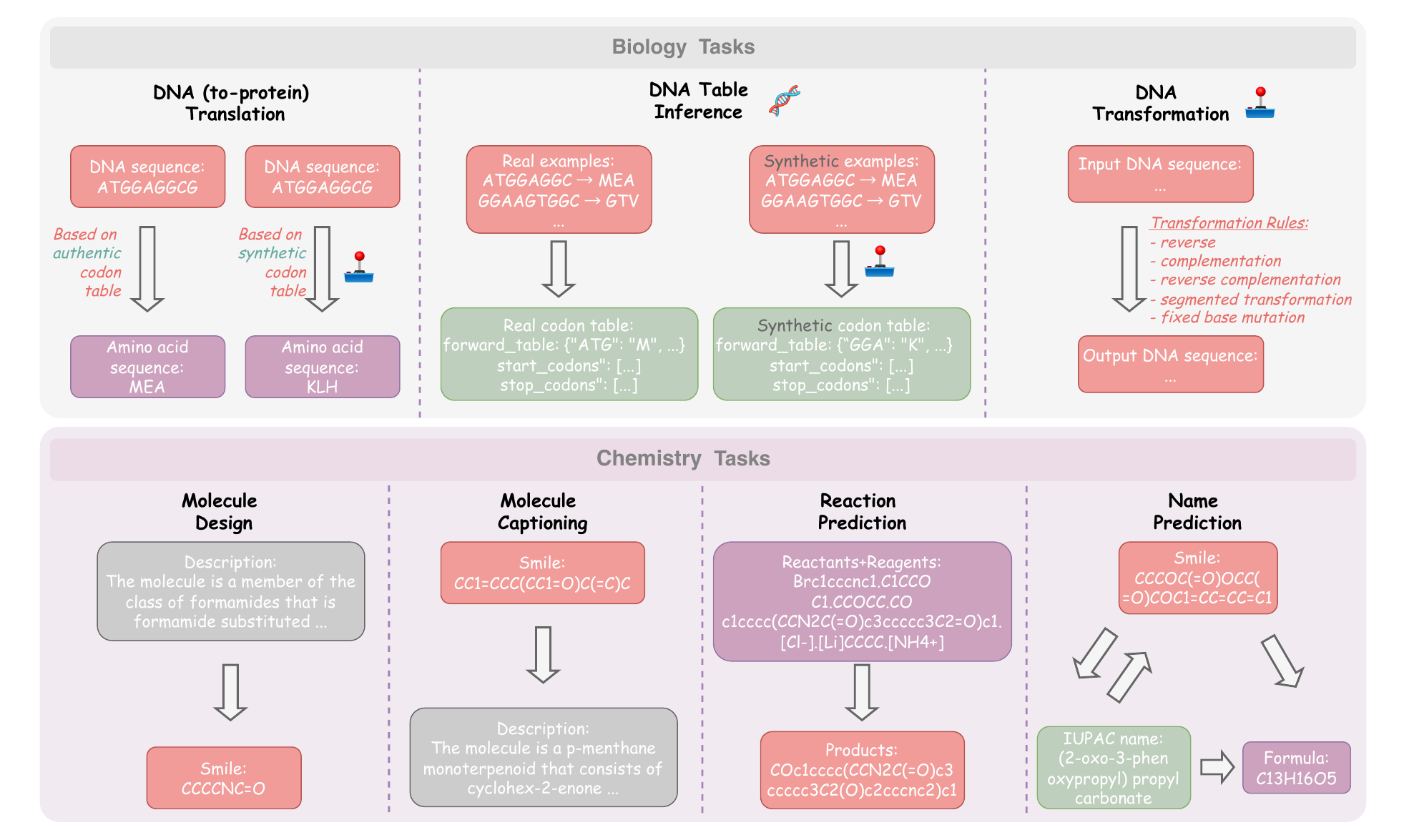}
    \caption{Our benchmark includes 7 tasks spanning two scientific disciplines: biology and chemistry. \includegraphics[height=1em]{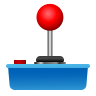} denotes tasks that adopt a synthetic configuration; \includegraphics[height=1em]{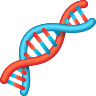} refers to tasks that involve only rule induction from examples, while others involve both induction and application to a new test input.}

	\label{fig:task_overview}
\end{figure*}

% \begin{figure*}[!t]
% 	\centering
% 	\includegraphics[width=\linewidth]{imgs/infer_strategies.pdf}
%     \caption{xxx}

% 	\label{fig:task_overview}
% \end{figure*}

\subsection{Task Overview}
\paragraph{Task 1: DNA Translation (Synthetic)} 
This task simulates the biological process of translating a DNA sequence into its corresponding amino acid sequence. The model is required to induce the codon-to-amino-acid mappings solely based on in-context learning (ICL) examples and apply the inferred mappings to translate a target DNA sequence. However, LLMs may have internalized the canonical genetic codon table as prior knowledge, enabling them to generate the correct amino acid sequence through memorization rather than genuine rule induction. To better assess the inductive reasoning capabilities of the model, we provide a synthetic alternative to the standard task design, by randomly assigning codon-to-amino-acid mappings.

\paragraph{Task 2: DNA Table Inference (Synthetic)} 
This task focuses explicitly on evaluating the model’s inductive ability by requiring it to recover the underlying codon table based solely on a set of DNA–amino acid sequence pairs. The model is asked to infer the translation rules and provide a fully structured codon table, including codon-to-amino acid mappings, start codons, and stop codons. We follow the same design as in Task 1, providing both standard and synthetic configurations.

\paragraph{Task 3: DNA Transformation} 
This task adopts a fully synthetic setup, with the goal of evaluating the model’s ability to infer transformation rules from ICL examples and to apply them correctly to unseen test sequences. Each ICL example consists of an input–output DNA sequence pair generated by applying one of several predefined transformations: sequence reversal, complementation, reverse complementation, segmented transformation, and fixed base mutation.

\paragraph{Task 4: Molecule Design} 
This task requires LLMs to generate molecular structures that satisfy a given textual description. The input is a natural language sentence (in English), and the output is the corresponding molecule represented in SMILES format.

\paragraph{Task 5: Molecule Captioning} 
This task is the inverse of Task 4, where the input is a molecular structure and the model is expected to generate a corresponding description or annotation in natural language.

\paragraph{Task 6: Reaction Prediction}
This task focuses on chemical reaction prediction. Given one or more reactants and reagents, the model is expected to predict the resulting product in the form of a SMILES string. 

\paragraph{Task 7: Name Prediction}
% This task involves converting between three types of chemical representations: SMILES, IUPAC names, and molecular formulas. SMILES encodes molecular structures as linear strings capturing atoms and connectivity, while IUPAC names are systematic, human-readable labels based on standardized nomenclature rules. Molecular formulas summarize atomic composition (e.g., \(\mathrm{C_2H_6O}\)) but lack structural information. Since a single formula can map to multiple SMILES due to isomerism, such mappings are ambiguous and less suitable for evaluation. We therefore focus on four conversion directions: SMILES to molecule formula translation~({\em smiles2formula}), SMILES to IUPAC name translation~({\em smiles2iupac}), IUPAC name to SMILES translation~({\em iupac2smiles)} and IUPAC name to
% molecule formula translation~({\em iupac2formula)}.
This task focuses on conversions between three common chemical representations: SMILES (linear structural encodings), IUPAC names (standardized nomenclature), and molecular formulas (atomic composition). We include four relatively unambiguous conversions: {\em smiles2formula}, {\em smiles2iupac}, {\em iupac2smiles}, and {\em iupac2formula}. 

\subsection{Data Collection}

\paragraph{Biology}
We derive source DNA sequences and their corresponding amino acid sequences from GenomicLLM\_GRCh38~\cite{grevsova2023genomic,Liu2024ExploringGL} for the standard task. For the synthetic task, we generate codon tables by randomizing every mapping except the start and stop codons, and translate inputs using these tables.

For DNA Transformation, we randomly sample DNA fragments from the training set as ICL examples and truncate them to a maximum length, and do the same for test sequences. The transformation type and base-pairing schemes are randomly sampled from a predefined set. These base-pairing schemes are designed manually to disrupt natural complementarity, increasing the inductive reasoning challenge. For all the tasks, we ensure that the ICL examples cover all the mappings used in the test example.

\paragraph{Chemistry}
ChemLLMBench~\cite{Guo2023WhatCL} is a chemistry-domian LLM benchmark comprising eight tasks. We select four tasks, corresponding to Task 4-7 in our work, which exhibit a relatively stronger emphasis on inductive reasoning capabilities. The Molecule Design and Captioning tasks are based on the ChEBI-20 dataset~\cite{edwards-etal-2022-translation}, pairing molecular SMILES with textual description. The Reaction Prediction task draws on the USPTO-MIT Mixed reaction dataset~\cite{Irwin2021ChemformerAP,Westerlund2024DoCD,westerlund2024constrained}, which contains information on reactants, reagents, and products in SMILES reaction format. The Name Prediction task is derived from PubChem~\cite{Kim2018PubChem2U}, which offers extensive mappings between SMILES strings and their corresponding standard chemical names, including both IUPAC names and molecular formulas.

\subsection{Metrics}
\label{sec:metrics}

\paragraph{Biology}
All three tasks are evaluated using accuracy as the primary metric, computed as the proportion of correctly predictions. 

% For DNA translation and transformation, accuracy is measured at the character level, with whitespace and stop codons removed. For table inference, accuracy is computed over the structural components of the codon table, including forward mappings, start codons, and stop codons, with unparsable predictions assigned a score of zero.

\paragraph{Chemistry}
For molecule design, we adopt eight metrics, including BLEU, Exact Match~\cite{Edwards2022TranslationBM}, and Levenshtein distance~\cite{Miller2009LevenshteinDI} for string-level consistency; validity for structural correctness; MACCS~\cite{ratcliff1988pattern}, RDK~\cite{rdkit}, and Morgan~\cite{Dash2023EvaluationOG} for structural similarity; and FCD~\cite{Preuer2018FrchetCD} for distributional similarity. For molecule captioning, we use BLEU, ROUGE, and METEOR to capture surface-level overlaps, but also introduce an LLM-as-a-Judge score (1–10 scale), with an emphasis on scientific accuracy, while also considering completeness and clarity.  For reaction prediction, we follow the Top-1 Accuracy metric and improve robustness by canonicalizing both predicted and reference SMILES using RDKit~\cite{rdkit} before comparison. Finally, for name prediction, we apply the same canonicalization for the~{\em iupac2smiles} task, and adopt Exact Match Accuracy for the other three tasks ({\em smiles2formula}, {\em smiles2iupac}, and {\em iupac2formula}).

\section{Evaluation}
% We conduct extensive experiments on various LLMs using our proposed benchmark. Additionally, we evaluated a range of inference strategies in order to assess the effectiveness of these methods for inductive reasoning.
\subsection{Models}
In order to provide a comprehensive assessment of the inductive reasoning capabilities of cost-optimized, flagship, and reasoning LLMs, we choose one representative model from each category, namely Claude-3.5-Haiku, GPT-4.1, and Gemini-2.5-Flash. Since our benchmark is integrated into the OpenCompass framework, it can be easily evaluated on any other LLM. To ensure consistency and encourage output diversity during repeated sampling, we set the temperature at $1.0$ for all experiments. For Gemini-2.5-Flash, we retain its default “thinking” configuration.
\subsection{Inference Strategies}
\begin{figure*}[!t]
	\centering
	\includegraphics[width=\linewidth]{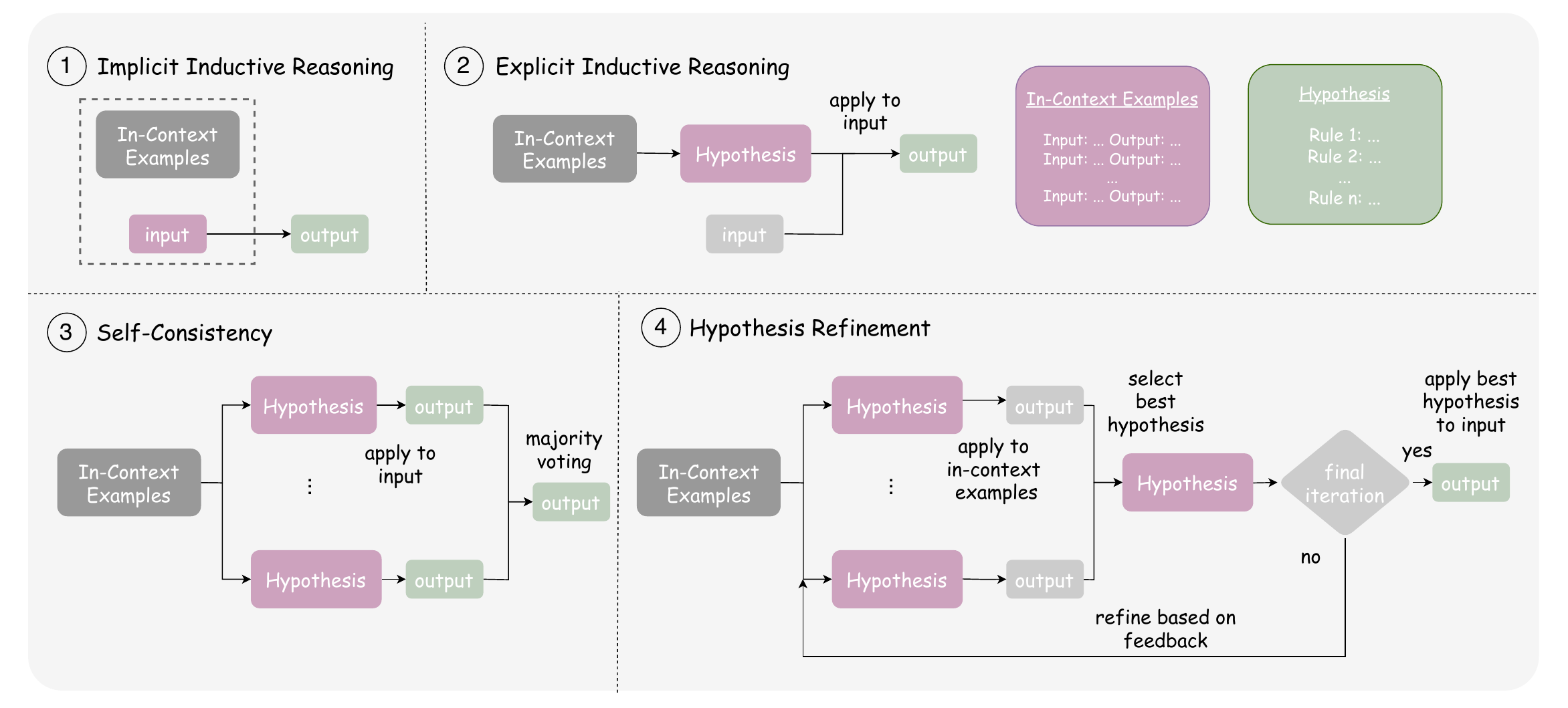}
    \caption{Comparison of four inference strategies: (1) Implicit induction - directly providing output; (2) Explicit induction - formulating clear hypotheses explicitly; (3) Self-consistency - using multiple reasoning paths to reach consensus; and (4) Hypothesis refinement - iteratively improving hypothesis on feedback.}

	\label{fig:task_overview}
\end{figure*}
We evaluate SIRBench-V1 on four commonly used inference strategies for inductive reasoning as illustrated in figure~\ref{fig:task_overview}. Explicit inductive reasoning serves as a baseline for advanced methods like self-consistency and hypothesis refinement, where the LLM needs to explicitly formulate and apply the hypotheses.
% Implicit inductive reasoning only requires the model to give the answer directly, whereas explicit inductive reasoning requires the model to provide induced rule.
% \comments{SC and HR are under explicit inductive reasoning, but with advanced strategies}

\paragraph{Implicit Inductive Reasoning.} We provide the LLM with ICL examples and ask the LLM to provide the final answer directly without explicitly stating the induced rules. This approach is the most straightforward way to perform inductive reasoning.

\paragraph{Explicit Inductive Reasoning.} We prompt the LLM to formulate a hypothesis based on the ICL examples. Then, we let the LLM apply the hypothesis to the given target question to obtain the final answer. This approach forces the LLM to perform the inductive reasoning process explicitly.

\paragraph{Self-Consistency.} For self-consistency ~\cite{wang2022self}, we sample multiple hypotheses (we use $n=5$) from the LLM and ask it to apply each of them to the target question, obtaining a corresponding answer from each hypothesis. A final answer is selected using majority voting performed by the LLM itself via prompting (see appendix \ref{sec:appendix}).

\paragraph{Hypothesis Refinement.} The hypothesis refinement method ~\cite{Qiu2023PhenomenalYP} follows a three-stage iterative process: hypothesis generation, selection, and refinement. 

Initially, we sample multiple hypotheses ($n=5$) based on the ICL examples, then evaluate them using one of the two approaches: (1) for code-executable tasks, we translate them into Python functions and execute them following~\citet{Qiu2023PhenomenalYP}, or (2) otherwise, we have the LLM apply each hypothesis directly. A task-specific evaluator scores each hypothesis's output.

Next, we generate a new set of hypotheses ($n=5$) by prompting (see appendix \ref{sec:appendix} for prompt) the LLM to refine the highest-scoring hypothesis based on feedback.

We repeat this select-and-refine loop up to $t=3$ iterations, stopping early if the hypothesis achieves a perfect score on ICL examples or performance degradation is detected.
We added the early stopping mechanism for performance degradation to prevent weaker models from degrading rule quality.

Finally, we apply the best resulting hypothesis to the target question to produce the answer.

\section{Results and Analysis}

% \comments{molecule design morgan sims is from 0 to 1. make it into percentage. make the task name abbreviated}

\subsection{Main Results}

\begin{table*}[t!]
  \centering
  \resizebox{\textwidth}{!}{
    \begin{tabular}{l  *{3}{c}  *{4}{c} c}
      \toprule
      \multirow{3}{*}{\textbf{Models}}
        & \multicolumn{3}{c}{\textbf{Biology}}
        & \multicolumn{4}{c}{\textbf{Chemistry}}
        & \multirow{3}{*}{\bf Avg.} \\
      \cmidrule(lr){2-4} \cmidrule(lr){5-8}
      % second header row: the individual tasks
      % & DNA Translation
      % & DNA Table Inference
      % & DNA Transformation
      % & Molecule Design
      % & Molecule Caption
      % & Reaction Prediction
      % & Name Prediction \\
      
      & \multicolumn{1}{c}{DNA} & \multicolumn{1}{c}{DNA Table} & \multicolumn{1}{c}{DNA} & \multicolumn{1}{c}{Molecule} & \multicolumn{1}{c}{Molecule} & \multicolumn{1}{c}{Reaction} & \multicolumn{1}{c}{Name} & \\
      & \multicolumn{1}{c}{Translation} & \multicolumn{1}{c}{Inference} & \multicolumn{1}{c}{Transformation} & \multicolumn{1}{c}{Design} & \multicolumn{1}{c}{Caption} & \multicolumn{1}{c}{Prediction} & \multicolumn{1}{c}{Prediction} \\
      
      \midrule
      
      % \addlinespace[0.5ex]  % small vertical gap, optional
      \multicolumn{9}{c}{\textbf{Implicit Inductive Reasoning}} \\
      % \addlinespace[0.5ex]  % small vertical gap, optional
      \midrule

      Claude-3.5-Haiku
        & 5.47 & 10.23 & 27.28 & 62.00 & 67.70 & \underline{44.44} & 3.57 & 31.53  \\
      GPT-4.1
        & 5.71 & 12.73 & \underline{31.37} & 75.00 & 66.30 & 22.22 & 13.51 & 32.41 \\
      Gemini-2.5-Flash
        & \textbf{11.72} & \textbf{32.06} & 30.42 & \textbf{85.00} & 63.30 & \textbf{54.17} & 30.00 & \textbf{43.81} \\

      \midrule
      % \addlinespace[1.0ex]  % small vertical gap, optional
      \multicolumn{9}{c}{\textbf{Explicit Inductive Reasoning}} \\
      % \addlinespace[0.5ex]  % small vertical gap, optional
      \midrule

      Claude-3.5-Haiku
        & 5.85 & 9.72 & 26.05 & 64.00 & 54.00 & 19.23 & 2.81 & 25.95 \\
      GPT-4.1
        & 5.31 & 12.13 & 28.73 & 69.00 & 59.00 & 17.86 & 6.09 & 28.30 \\
      Gemini-2.5-Flash
        & 9.14 & 23.34 & 28.66 & 77.00 & \underline{67.70} & 34.78 & 30.00 & 38.66 \\

      \midrule
      % \addlinespace[1.0ex]  % small vertical gap, optional
      \multicolumn{9}{c}{\textbf{Self-Consistency~\cite{wang2022self}}} \\
      % \addlinespace[0.5ex]  % small vertical gap, optional
      \midrule

      Claude-3.5-Haiku
        & 5.11 & 10.00 & 26.34 & 66.00 & 69.70 & 20.83 & 0.83 & 28.40 \\
      GPT-4.1
        & 5.96 & 13.19 & 30.81 & 72.00 & 65.70 & 25.00 & 9.58 & 31.75 \\
      Gemini-2.5-Flash
        & 9.15 & 24.84 & 30.4 & \underline{80.00} & \textbf{70.00} & 39.29 & \textbf{40.13} & \underline{41.97} \\

      \midrule
      % \addlinespace[1.0ex]  % small vertical gap, optional
      \multicolumn{9}{c}{\textbf{Hypothesis Refinement~\cite{Qiu2023PhenomenalYP}}} \\
      % \addlinespace[0.5ex]  % small vertical gap, optional
      \midrule

      Claude-3.5-Haiku
        & 5.79 & 10.02 & 30.05 & 73.00 & \textbf{72.70} & 28.00 & 1.88 & 31.63 \\
      GPT-4.1
        & 5.62 & 14.57 & \textbf{35.56} & 67.00 & 66.30 & 32.14 & 11.59 & 33.25 \\
      Gemini-2.5-Flash
        & \underline{10.60} & \underline{28.55} & 30.37 & 72.00 & 65.70 & 32.14 & \underline{34.07} & 39.06 \\

      \bottomrule
    \end{tabular}
  }
  \caption{Performance of Claude-3.5-Haiku, GPT-4.1, and Gemini-2.5-Flash on SIRBench-V1 using four inference strategies. All scores report accuracy (\%), except Molecule Design (Morgan similarity rescaled to 0-100). Molecule Caption reports the accuracy from LLM-as-judge. Synthetic versions were used for DNA Translation and DNA Table Inference tasks.}
  \label{tab:main-results}
\end{table*}

Table \ref{tab:main-results} reveals consistently low performance across most tasks, highlighting the limitations of current LLMs in scientific inductive reasoning tasks beyond mathematical equations. Among the evaluated models, Gemini-2.5-Flash demonstrates superior performance in computationally intensive tasks while exhibiting comparable results to other models in conceptually oriented tasks such as Molecule Caption. Additionally, larger flagship models perform better than cost-optimized models.

We observe that LLMs struggle with explicit inductive reasoning (i.e., proposing effective rules and applying them to novel inputs), as shown by the performance drop from implicit to explicit inductive reasoning. Self-consistency helps alleviate this shortcoming by sampling multiple diverse reasoning paths and marginalizing across them, thereby enhancing the robustness of the explicit inductive reasoning process. The hypothesis refinement strategy further improves the performance, as it selects the best rule from multiple sampled hypothesis and revises the rule at each iteration. However, we find that the advantage of hypothesis refinement over implicit inductive reasoning varies inconsistently across tasks and models.

\begin{table*}[t!]
  \centering
  \resizebox{\textwidth}{!}{
    \begin{tabular}{l  *{3}{c}  *{4}{c} c}
      \toprule
      \multirow{3}{*}{\textbf{Models}}
        & \multicolumn{3}{c}{\textbf{Biology}}
        & \multicolumn{4}{c}{\textbf{Chemistry}}
        & \multirow{3}{*}{\bf Avg.} \\
      \cmidrule(lr){2-4} \cmidrule(lr){5-8}
      
      & \multicolumn{1}{c}{DNA} & \multicolumn{1}{c}{DNA Table} & \multicolumn{1}{c}{DNA} & \multicolumn{1}{c}{Molecule} & \multicolumn{1}{c}{Molecule} & \multicolumn{1}{c}{Reaction} & \multicolumn{1}{c}{Name} & \\
      & \multicolumn{1}{c}{Translation} & \multicolumn{1}{c}{Inference} & \multicolumn{1}{c}{Transformation} & \multicolumn{1}{c}{Design} & \multicolumn{1}{c}{Caption} & \multicolumn{1}{c}{Prediction} & \multicolumn{1}{c}{Prediction} \\
      
      \midrule

      Qwen3-8B (with thinking)
        & 0.20 & 4.88 & 3.24 & 59.00 & 52.67 & 3.33 & 1.67 & 17.00 \\
      Qwen3-8B (without thinking)
        & 6.30 & 7.06 & 27.19 & 50.00 & 49.67 & 0.00 & 0.00 & 20.03 \\
      Deepseek-V3-0324
        & 7.21 & 12.24 & 28.81 & 75.00 & 64.00 & 30.00 & 14.17 & 33.06 \\
      
      \bottomrule
    \end{tabular}
  }
  \caption{Performance of Qwen3-8B and Deepseek-V3-0324 on SIRBench-V1  under the \textbf{Implicit Inductive Reasoning} setting. Scores are accuracy (\%) except Molecule Design (Morgan similarity, 0-100 scale) and Molecule Caption (LLM-as-judge accuracy). Synthetic versions used for DNA tasks.}
  \label{tab:opensource-implicit}
\end{table*}

To validate our findings across more LLMs, we evaluated additional open-source models under implicit inductive reasoning, as shown in Table \ref{tab:opensource-implicit}. Deepseek-V3-0324 performs comparably with GPT-4.1 across most tasks, while Qwen3-8B with thinking generates extremely long chain-of-thought reasoning for biology tasks, often exceeding its recommended 32K max output length without completing the reasoning process, demonstrating that long chain-of-thought is not effective on the biology tasks. These results reinforce our findings on the fundamental limitation of current LLMs in scientific inductive reasoning. Additionally, current inductive reasoning methods remain inadequate for scientific inductive reasoning tasks beyond mathematical equations.

% This raises the question on the effectiveness of current inductive reasoning methods.
% , raising the question whether LLMs can truly perform effective inductive reasoning
% Compared to implicit, Hypothesis refinement shows mixed results across tasks, with notable improvements in biology tasks. This suggests that iterative hypothesis refinement on in-context examples benefits tasks with well-defined patterns. However, it offers limited advantages for tasks demanding deep conceptual understanding. Therefore, current inductive reasoning methods only have limited improvements.

% IDEA: implicit better than explicit
% under explicit, SC and HR improves performance
% explicit bad, consistent w prv works
% By observing the gap between DNA Translation task and DNA Table inference task which involve the same type of rules, we see that LLMs also struggle to . 

\subsection{Effect of Length}
\begin{figure}
    \centering
    \includegraphics[width=\columnwidth]{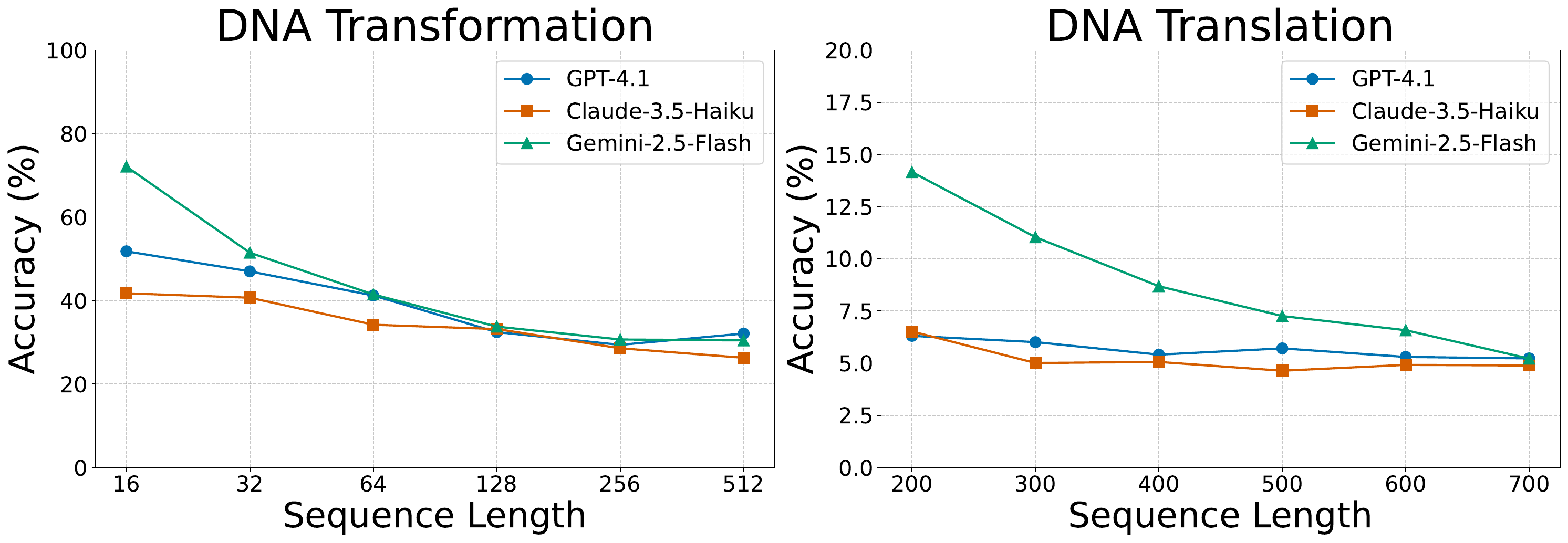}
    \caption{Effect of Sequence Length in Transformation and DNA Translation tasks.}
    \label{fig:length-comparison}
\end{figure}
% one key insight here is that when length is very small, then it is similar to previous synthetic benchmarks
% but for very long inputs, previous synthetic benchmarks don't cover
% this is not the same as real world application where the inputs are long, does not match real-life scenarios
% we see that models can perform well on short inputs, but not on long ones
% start from very small length, intervals of 50 for table inference and translation
Being able to perform inductive reasoning on a long context is fundamental. We evaluated the LLMs on DNA transformation and DNA translation tasks with varying sequence length configurations. The DNA transformation task demands the comprehension of the entire sequence (e.g., identifying reversals), while the DNA translation task requires observation of local patterns. As shown in figure \ref{fig:length-comparison}, for DNA transformation, we found that the LLMs achieve relatively strong performance on shorter sequences but exhibits a significant performance decline as sequence length increases. For DNA translation, GPT-4.1 and Claude-3.5-Haiku show minimal decrease with longer sequences only because they struggle with this task at shorter lengths. The results indicate that current LLMs are effective at inducing pattern only within limited input lengths. This limitation reflects the broader challenge of developing robust inductive reasoning capabilities that can handle long context.

\subsection{Effect of Number of Shots}

\begin{figure}
    \centering
    \includegraphics[width=\columnwidth]{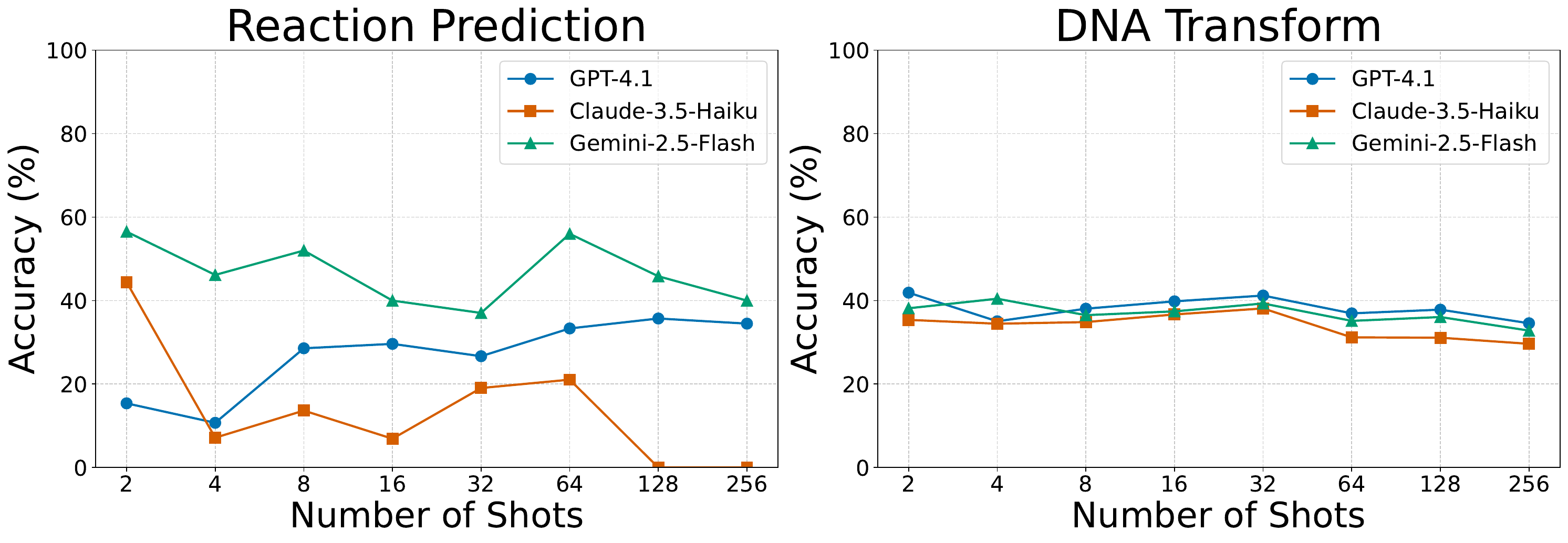}
    \caption{Effect of Number of Shots in Reaction Prediction and DNA Transformation tasks.}
    \label{fig:shots-comparison}
\end{figure}
We examine the effect of the number of shots on accuracy in one representative task each from the domains of biology and chemistry. Figure~\ref{fig:shots-comparison} shows that increasing the number of shots has varying effects on different models. In reaction prediction task, GPT-4.1 exhibits an upward trend, showing that it benefits from additional shots. In contrast, Claude-3.5-Haiku shows performance degradation, likely due to limitations in its context processing capability. Gemini-2.5-Flash does not show any clear upward or downward trend with as shot increases. For DNA transformation, all the models exhibit consistent performance, implying that additional examples provide limited benefit.

\subsection{Many-Short-Shot vs. Long-Few-Shot}
\begin{table}[t]
\centering
\resizebox{\linewidth}{!}{
\begin{tabular}{l c c}
\toprule
\textbf{Model} & \textbf{Many-Short-Shot} & \textbf{Few-Long-Short} \\
\midrule
Claude-3.5-Haiku & 31.19 & 15.63 \\
GPT-4.1 & 36.94 & 25.64 \\
Gemini-2.5-Flash & 35.14 & 24.47 \\
\bottomrule
\end{tabular}}
\caption{Performance comparison in many-short-shot versus long-few-shot settings on the DNA Translation task. The many-short-shot setting uses 64 shots with sequence length 100, while the few-long-shot setting uses 4 shots with sequence length 1600.}
\label{tab:shot-comparison}
\end{table}

Unlike previous studies that only explore increasing the number of relatively short examples~\cite{Yan2025MIRBenchBL}, we also explore the inductive reasoning capabilities of LLMs on few long examples. The latter paradigm adheres more to real-world applications, where it is difficult to obtain numerous examples for long input tasks. Our comparative analysis in table \ref{tab:shot-comparison} across both scenarios while maintaining the total input length demonstrates that LLMs perform worse with few long examples. This finding highlights a critical area for the advancement of LLM inductive reasoning ability.
% the two configurations are (64 shots, 100 seq len) and (4 shots, seq len 1600) respectively

\subsection{Task Difficulty Analysis}
Reasoning ability is not only reflected in overall accuracy but also in performance across difficulty levels. We analyzed two representative tasks, one from biology and one from chemistry, under Implicit Inductive Reasoning. Test instances were categorized into Easy, Medium, Hard, with 100 samples each. The DNA Translation samples were grouped by input sequence length, with ranges of 100-300 for Easy, 300-500 for Medium, and 500-700 for Hard, while the Molecule Design samples were classified by molecular complexity using RDKit based on structural features. As shown in both Table \ref{dnatable} and Table \ref{moletable}, model performance exhibits a clear downward trend from easy to hard samples, suggesting that difficulty-based categorization offers a straightforward way to assess robustness while also enabling a more fine-grained evaluation of reasoning abilities across domains.

% \begin{table}[htbp]
%   \centering

%   \resizebox{\columnwidth}{!}{%
%   \scriptsize
%     \begin{tabular}{|c|c|c|c|}
%     \toprule
%     \textbf{Difficulty Level} & \textbf{GPT-4.1} & \textbf{Claude} & \textbf{Gemini} \\
%     \midrule
%     \textbf{Easy} & 6.16  & 5.77  & 12.6 \\
%     \midrule
%     \textbf{Medium} & 5.56  & 4.85  & 7.98 \\
%     \midrule
%     \textbf{Hard} & 5.27  & 4.91  & 5.9 \\
%     \bottomrule
%     \end{tabular}%
%     }
%     \caption{Performance of LLMs on the DNA Translation task by difficulty level.}
%   \label{dnatable}%
% \end{table}%

\begin{table}[htbp]
  \centering
  \resizebox{\columnwidth}{!}{%
    \begin{tabular}{ll ccc}
    \toprule
    \multicolumn{2}{c}{\textbf{Difficulty Level}} & \textbf{GPT-4.1} & \textbf{Claude} & \textbf{Gemini} \\
    \midrule
    \textbf{Easy} & accuracy & 6.16  & 5.77  & 12.6 \\
    \midrule
    \textbf{Medium} & accuracy & 5.56  & 4.85  & 7.98 \\
    \midrule
    \textbf{Hard} & accuracy & 5.27  & 4.91  & 5.9 \\
    \bottomrule
    \end{tabular}%
    }
    \caption{Performance of LLMs on the DNA Translation task by difficulty level.}
  \label{dnatable}%
\end{table}%

\begin{table}[htbp]
  \centering
  \resizebox{\columnwidth}{!}{%
    \begin{tabular}{ll ccc}
    \toprule
    \multicolumn{2}{c}{\textbf{Difficulty Level}} & \textbf{GPT-4.1} & \textbf{Claude} & \textbf{Gemini} \\
    \midrule
    \multirow{3}{*}{\textbf{Easy}} 
          & validity & 0.94  & 0.67  & 0.94 \\
\cmidrule{2-5}          
          & morgan\_sims & 0.67  & 0.39  & 0.89 \\
\cmidrule{2-5}          
          & fcd (↓) & 2.66  & 9.82  & 1.15 \\
    \midrule
    \multirow{3}{*}{\textbf{Medium}} 
          & validity & 0.92  & 0.64  & 0.88 \\
\cmidrule{2-5}          
          & morgan\_sims & 0.55  & 0.29  & 0.78 \\
\cmidrule{2-5}          
          & fcd (↓) & 7.77  & 21.08 & 4.73 \\
    \midrule
    \multirow{3}{*}{\textbf{Hard}} 
          & validity & 0.74  & 0.59  & 0.41 \\
\cmidrule{2-5}          
          & morgan\_sims & 0.46  & 0.21  & 0.6 \\
\cmidrule{2-5}          
          & fcd (↓) & 19.85 & 29.86 & 22.24 \\
    \bottomrule
    \end{tabular}%
    }
    \caption{Performance of LLMs on the Molecule Design task by difficulty level.}
  \label{moletable}%
\end{table}%

\subsection{Counterfactual Evaluation}
\begin{table}[ht]
    \centering
    \resizebox{\linewidth}{!}{
    \begin{tabular}{l lr lr}
    \toprule
    \multirow{2}{*}{\textbf{Model}}
      & \multicolumn{2}{c}{\textbf{DNA Translation}}
      & \multicolumn{2}{c}{\textbf{DNA Table Inf.}} \\
    \cmidrule(lr){2-3} \cmidrule(lr){4-5}
      & \textbf{Aut.} & \textbf{Syn. ($\Delta$)}
      & \textbf{Aut.} & \textbf{Syn. ($\Delta$)} \\ 
    \midrule
    Claude-3.5-Haiku
      & 21.95
      & 5.47 {\scriptsize($-16.48$)}
      & 68.50
      & 10.23 {\scriptsize($-58.27$)} \\
    GPT-4.1
      & 21.24
      & 5.71 {\scriptsize($-15.53$)}
      & 81.84
      & 12.73 {\scriptsize($-69.11$)} \\
    Gemini-2.5-Flash
      & 30.64
      & 11.72 {\scriptsize($-18.92$)}
      & 87.09
      & 32.06 {\scriptsize($-55.03$)} \\
    \bottomrule
    \end{tabular}
    }
    \caption{Performance comparison between authentic and synthetic versions of chosen tasks. $\Delta$ represents the performance gap, calculated as the score on synthetic tasks minus the score on authentic tasks.}
    \label{tab:counterfactual_performance}
\end{table}

To investigate whether LLMs perform true inductive reasoning, we compare their performance on original and synthetic settings of DNA Translation and Table Inference. As illustrated in Table \ref{tab:counterfactual_performance}, all three models suffer a dramatic performance decline in synthetic tasks, suggesting that higher performance in authentic versions stems from the memorization of standard mappings rather than genuine inductive reasoning capabilities.

Among the evaluated models, Gemini-2.5-Flash maintains the highest performance on both original and synthetic versions of the tasks. This suggests that reasoning models have better capability to identify rules beyond the constraints of memorized knowledge than non-reasoning models. However, its absolute score in synthetic tasks remains low. Overall, these results indicate that current LLMs are fundamentally limited in their ability to perform genuine inductive reasoning. In the context of scientific discovery, LLMs need to recognize novel patterns rather than just retrieve existing knowledge. Therefore, our findings highlight the need to distinguish inductive reasoning from retrieval to advance the ability of LLMs for scientific discovery.

\section{Conclusion}
In this paper, we introduce SIRBench-V1, a benchmark that includes Chemistry and Biology subtasks, to evaluate the scientific inductive reasoning of LLMs on tasks beyond mathematical equation. We evaluated different LLMs using commonly used reasoning strategies on our proposed benchmark. We found that current leading LLMs obtain low performance on our benchmark and that using sophisticated strategies provide minimal benefits. Additionally, we point out limitations of LLMs in performing inductive reasoning on longer context lengths, few-long-shot settings, and counterfactual rules. The experimental results will provide valuable insights for future studies on LLM-driven scientific discovery.
% task what we do
% conclusion
% implications

\section{Limitations}
In this work, we take the first step toward incorporating scientific scenarios into the design of the LLM-Based Inductive Reasoning Beyond Equations and introduce a new dataset for evaluation. However, the SIRBench-V1 is limited to chemistry and biology domains. As a next step, we plan to invite domain experts in these areas to review and refine both our benchmark and evaluation protocol. In the future, we aim to expand the benchmark to cover a broader range of scientific disciplines.

% \highlight{
\section*{Acknowledgement}
This work is supported by the AI9Stars community, the Fundamental Research Funds for the Central Universities, and the Research Funds of Beijing Language and Culture University (25YCX118). We also thank the anonymous reviewers in ACL Rolling Review May 2025. Their insightful feedback and suggestions are significant to refining and improving our work.

\bibliography{custom}
\clearpage
\appendix

\section{Additional Details on SIRBench-V1}
\subsection{Dataset Configurations}
We curate 7 tasks in total. Considering that multiple metrics provide robust assessment, for chemistry tasks, we evaluate Molecule Design, Molecule Captioning and Reaction Prediction with 30 examples each. For Name Prediction, we sample 30 examples for each type of transformation (including {\em smiles2formula}, {\em smiles2iupac}, {\em iupac2smiles}, and {\em iupac2formula}). Since biology tasks rely solely on accuracy, we increase the number of examples to 100 for each biology task to ensure more stable evaluation, including DNA Translation, DNA Translation (Synthetic), DNA Table Inference, DNA Table Inference (Synthetic) and DNA Transformation. All experiments are conducted under 5-shot setting, unless otherwise stated. However, since our benchmark has various configurations and supports synthetic data generation for some subtasks, the actual number of items can be configurable. 

In our main results, we use the following configurations. For DNA Translation, we uniformly sample across sequence length 200 to 450 since the effective DNA sequences in the dataset starts from length 200. While data are available for longer sequences, only sample until 450 because they are too challenging for most models. For DNA Transformation, we set the sequence length to 300, which is a reasonably challenging level.

\subsection{Examples of Transformation Types in DNA Transformation Task}
The transformation types include: \textbf{1) Sequence reversal}: reversing the order of the entire sequence (e.g., AGCT → TCGA); \textbf{2) Complementation}: replacing each base according to a substitution rule (e.g., AGCT → TCGA, using A$\leftrightarrow$T, C$\leftrightarrow$G or a randomized complement map); \textbf{3) Reverse complementation}: performing complementation followed by reversal (e.g., AGCT → AGCT); \textbf{4) Segmented transformation}: transforming fixed-length segments after a fixed stride (e.g., AGCTTAGCGT → AGCTTGACGT, reversing 2 bases every 3 bases); \textbf{5) Fixed base mutation}: replacing specific bases with new ones (e.g., AGCT → GGTT, where A→G and C→T).
\section{Explicit Inductive Reasoning Analysis}
\subsection{Hypothesis Quality and Refinement}
\begin{table}[ht]
    \centering
    \resizebox{\columnwidth}{!}{%
    \begin{tabular}{llccc}
    \toprule
    \textbf{Task} & \textbf{Model} & \textbf{Initial} & \textbf{Final} & \textbf{Test} \\
    \midrule
    \multirow{3}{*}{DNA Translation} & Claude-3.5-Haiku & 3.87 & 6.52 & 5.79 \\
    & GPT-4.1 & 9.15 & 11.37 & 5.62 \\
    & Gemini-2.5-Flash & 24.37 & 30.57 & 10.60 \\
    \midrule
    \multirow{3}{*}{Molecule Design} & Claude-3.5-Haiku & 0.67 & 0.71 & 0.73 \\
    & GPT-4.1 & 0.77 & 0.82 & 0.67 \\
    & Gemini-2.5-Flash & 0.92 & 0.97 & 0.72 \\
    \bottomrule
    \end{tabular}
    }
    \caption{Comparison of initial and final hypothesis quality scores on in‑context examples (ICE) alongside corresponding test performance of final hypothesis for various models across DNA Translation (Synth) and Molecule Design tasks. Morgan similarity (scale of 0 to 1) is reported for the Molecule design task.}
    \label{tab:hypothesis-quality}
\end{table}

In order to provide a more thorough analysis, we show the computed evaluation score of the generated hypotheses on ICL examples during hypothesis refinement in table~\ref{tab:hypothesis-quality}. For the initial evaluation scores, we report the average score of the best hypothesis generated by the model prior to any refinement. This also serves as an approximate upper bound of the evaluation scores for hypotheses generated by explicit inductive reasoning and self-consistency. We notice that for DNA Translation task, these rules obtained low accuracy on ICL examples. The scores increase at the end of the refinement process, but still remain low. This shows the limitation in inductive reasoning capability of current LLMs. Furthermore, although hypothesis refinement enhances the consistency of rules with ICL examples, we observe that in some cases there remains a substantial gap between performance on ICL examples and generalization to test examples. Future work may focus on enhancing the hypothesis proposal capabilities of LLMs to improve both accuracy on ICL examples and generalization to novel scenarios.

We also provided examples of hypothesis generated by GPT-4.1 which can be found table \ref{tab:dna-hypotheses} and \ref{tab:chem-hypotheses}.

% Table~\ref{tab:hypothesis-quality} reveals that while all models show improvement in hypothesis quality on ICL examples after refinement, this improvement does not translate to better generalization in most cases. This suggests an overfitting where the LLM refines the hypothesis to better fit the given ICLexamples but does not account for the generalizability of the rule. While it is crucial that the hypothesis should account for all the ICL examples, the generalizability of the rules should also be maintained. Future research directions can explore methods that strike a balance between fitting ICL examples and extracting broadly applicable rules, thereby enhancing the model’s helpfulness for scientific discovery.
% overfitting, but less obvious for smaller models
% molecule design, hypothesis refinement does is overfitting. we need ineherent capabilities.
% task 1 suggest that will the hypothesis proposed by llm looks sound, they are not factually accurate

% We also provided examples in the table below

\begin{table*}[ht]
\centering
\resizebox{\textwidth}{!}{
\begin{tabular}{p{0.18\textwidth} p{0.78\textwidth}}
\toprule
\textbf{Task} & \textbf{Hypothesis} \\
\midrule
DNA Translation &
Rule 1: Read the DNA from 5'->3' in consecutive codons (3 bp) starting at the first base.  
Rule 2: Each codon uniquely maps to one amino acid per the in‐context examples.  
Rule 3: Inferred codon table: ATG:M, ATT:M, ATA:N, ATC:Q, GTT:S, GTC:V, GTA:C, GTG:V, TTT:F, TTC:F, TTA:W, TTG:W, CTT:M, CTC:D, CTA:R, CTG:G, TCT:S, TCC:S, TCA:H, TCG:S, CCT:P, CCC:N, CCA:N, CCG:Y, ACT:P, ACC:N, ACA:N, ACG:W, GCT:A, GCC:N, GCA:N, GCG:F, TAT:Y, TAC:P, TAA:*, TAG:*, CAT:Q, CAC:Y, CAA:D, CAG:V, AAT:F, AAC:G, AAA:K, AAG:N, GAT:D, GAC:G, GAA:E, GAG:G, TGT:P, TGC:L, TGA:*, TGG:W, CGT:R, CGC:C, CGA:M, CGG:H, AGT:S, AGC:F, AGA:G, AGG:S, GGT:G, GGC:W, GGA:V, GGG:Q.  
Rule 4: Translate sequentially until a stop (TAA, TAG, TGA → “*”).  
Rule 5: Output one‐letter amino acid string.  
Rule 6: Stop at and include first “*”.  
Rule 7: Unseen codons should not be assumed. \\
\midrule
DNA Table Inference &
Rule 1: DNA–protein pairs align codons (3 bp) to amino acids.  
Rule 2: Segment DNA into triplets from 5' and align to protein until “*” or end.  
Rule 3: Codons aligned to “*” are stop codons.  
Rule 4: First‐codon→‘M’ pairs are start codons.  
Rule 5: Aggregate across examples; record all observed mappings.  
Rule 6: Include only codons seen.  
Rule 7: Build forward\_table from all mappings, excluding stops.  
Rule 8: start\_codons = all first codons mapped to ‘M’.  
Rule 9: stop\_codons = all codons aligned to ‘*’.  
Rule 10: Amino acids are single‐letter codes including “*.” \\
\midrule
DNA Transform &
Rule 1: Split input into 7‑nt segments from 5'; last segment may be shorter.  
Rule 2: Reverse each 7‑nt segment.  
Rule 3: Concatenate reversed segments to form output. \\
\bottomrule
\end{tabular}}
\caption{Hypotheses Generated by GPT-4.1 for the DNA tasks}
\label{tab:dna-hypotheses}
\end{table*}

\begin{table*}[ht]
\centering
\resizebox{\textwidth}{!}{
\begin{tabular}{p{0.18\textwidth} p{0.78\textwidth}}
\toprule
\textbf{Task} & \textbf{Hypothesis} \\
\midrule
Molecule Design &
Rule 1: Identify required functional groups (e.g., diamine, aldehyde, etc.).  
Rule 2: Map biological role to known scaffolds (e.g., antineoplastic → stilbene).  
Rule 3: Choose core heterocycle per “derives from” (e.g., triazine).  
Rule 4: Decorate core with substituents to satisfy function and activity.  
Rule 5: Respect stereochemistry (e.g., [C@H] per natural enantiomer).  
Rule 6: For natural products, replicate known SMILES closely.  
Rule 7: Attach alkyl/aryl groups at correct positions.  
Rule 8: Output valid SMILES with rings, heteroatoms, charges. \\
\midrule
Molecule Caption &
Rule 1: Identify core ergot alkaloid and name (e.g., ergotaman).  
Rule 2: Describe substituents and positions (e.g., 12'‑hydroxy).  
Rule 3: Note stereochemistry if differentiating isomers.  
Rule 4: Mention salts/derivatives (e.g., methanesulfonic acid salt).  
Rule 5: State biological origin or role if recognizable.  
Rule 6: Use “derives from” for parent relationships.  
Rule 7: Note naming conventions or historical context if relevant.  
Rule 8: Separate distinct features into clear sentences. \\
\midrule
Reaction Prediction &
Rule 1: Target N‑heterocycle fused to benzene undergoes nucleophilic attack.  
Rule 2: Organometallics ([Li]CCCC, [H–]) add to carbonyl or halide.  
Rule 3: Bases ([NH$_4^+$], [OH–]) deprotonate or hydrolyze esters → amides/acids.  
Rule 4: Leaving groups replaced by nucleophiles forming C–X or C–C.  
Rule 5: Ester + nucleophile -> amide/ether.  
Rule 6: Most nucleophilic reagent reacts with most electrophilic center.  
Rule 7: Ignore spectator ions in final product.  
Rule 8: Grignard addition -> alcohol at addition site.  
Rule 9: Reductions ([H–]) convert carbonyls → alcohols/amines.  
Rule 10: On heteroaryl halide, nucleophile replaces halide on ring.  
Rule 11: Ethers/amides attach to aromatic systems via substitution/acylation.  
Rule 12: With both esters and amines, amide formation is preferred. \\
\midrule
Name Prediction &
Rule 1: Count all C atoms (including branches/rings).  
Rule 2: Count H via implicit valence rules.  
Rule 3: Count N, O, S, Si, halogens from SMILES.  
Rule 4: Include implicit Hs in aromatic rings per standard.  
Rule 5: Integrate substituent atoms without double‐counting.  
Rule 6: Adjust H count for double/triple bonds.  
Rule 7: Write formula as C, H, then others alphabetically.  
Rule 8: Expand grouped atoms (e.g., O[Si](C)(C)C).  
Rule 9: Sum counts; check branching consistency.  
Rule 10: Format as [Element][count]… (e.g., C6H6O). \\
\bottomrule
\end{tabular}}
\caption{Hypotheses Generated by GPT-4.1 for the Chemistry tasks}
\label{tab:chem-hypotheses}
\end{table*}

\subsection{Misalignment of Advanced Reasoning Strategies }
As shown in Table \ref{tab:main-results}, the performance of LLMs does not consistently improve with the application of more fine-grained reasoning strategies. In some cases, advanced strategies even reduce performance. To investigate this phenomenon, we analyzed the recorded reasoning traces, focusing on chemistry-related tasks. In the molecule captioning task, Self-Consistency occasionally produced lower scores than the Implicit Inductive Reasoning baseline. While this strategy generates multiple hypotheses and applies them to derive answers, the resulting outputs were often fragmented or overly technical. For example, instead of producing full descriptive captions, as required by the task, the model frequently produced structural abbreviations or linkage names such as beta-D-Galp (1→4) beta-D-GlcpNAc (which are often part of the rule representations extracted by the model), omitting information about overall structure or functional roles. This indicates a misalignment between rule-based derivations and the task's requirement for holistic descriptions. In the reaction prediction task, Hypothesis Refinement also failed to deliver consistent improvements. Our analysis suggests that this was due to refined rules were not always effectively applied to the examples, and the selection of the "best" hypothesis depended solely on an automatic evaluator of prediction accuracy, which does not necessarily capture scientific plausibility.

Overall, these results suggest that the limitations of advanced reasoning strategies stem less from insufficient domain knowledge in base models than from structural mismatches between the strategies and the nuanced demands of the tasks.

\section{Experiment Details}
\label{sec:appendix}

\label{sec:appendix}

\subsection{Implementation Details}
We run our experiments using API-based closed-source models, specifically claude-3-5-haiku-20241022, gpt-4.1-2025-04-14, and gemini-2.5-flash-preview-04-17. We implement our inference strategies in the OpenCompass framework. This allows us to perform inference in parallel at high rates. The explicit inductive reasoning is implemented via one-pass decoding, generating the hypothesis and applying it to the test example in one API call. Self-consistency is implemented by sampling multiple times using the same process as explicit inductive reasoning. For hypothesis refinement, we sample the hypothesis using the same general prompt in all tasks, except for DNA Translation where we ask the model to provide the specific codon-to-amino acid so that the hypothesis can be properly refined. For tasks in which the hypothesis can be translated into Python code, we prompt an LLM to generate the code. Otherwise, we prompt the LLM to apply a hypothesis to all in-context example inputs and do this to all the generated hypothesis. We used AI assistants to polish some of the text in this paper.

\subsection{Prompts}
\paragraph{Molecule Captioning}
As discussed in Section\ref{sec:metrics}, molecule captioning is an open-ended generation task, for which existing evaluations rely primarily on surface-level matching. To address this limitation, we design a dedicated prompt with fine-grained scoring criteria and employ an LLM to serve as the evaluator.

\paragraph{One-pass Self-Consistency}
To reduce the number of API calls and improve the efficiency of self-consistency, we design the prompt so that the model performs both rule induction and application to the test input within a single invocation.

\paragraph{Universal Majority Voting with Self-Consistency}
Given that the outputs of the chemistry and biology tasks in SIRBench-V1 are typically long and semantically complicated, basic majority voting mechanism often fails to identify a representative response, thereby diminishing the effectiveness of self-consistency. To address this, we adopt the universal self-consistency strategy\cite{Chen2023UniversalSF}, selecting the most semantically consistent response to form the final answer.

\paragraph{Hypothesis Refinement}
We provide the main prompts used in the hypothesis refinement process, including Hypothesis Induction, Hypothesis Application, Hypothesis Refinement, and Final Hypothesis Application.

\begin{figure*}[!htb]
\begin{tcolorbox}[
  colback=white,
  colframe=black,
  arc=2mm,
  boxrule=0.3pt,
  % breakable,
  % enforce breakable,
  % enhanced jigsaw,
  width=\textwidth,
]

\textbf{LLM-as-Judge Evaluation of Molecule Captioning: }

You are an expert molecular biologist.

Below is a SMILES string representing a molecule:  
\texttt{\{smiles\}}

Here is a reference description of the molecule:  
\texttt{\{gt\}}

Here is a predicted description of the same molecule:  
\texttt{\{pred\}}

Your task is to evaluate the \textbf{predicted} description \textbf{only} based on its scientific quality compared to the reference.

You must assign a \textbf{score from 1 to 10} based on the following criteria:

\begin{itemize}
  \item \textbf{Score 10}: Nearly perfect — scientifically precise, complete, and fluent. Matches all key aspects of the reference (e.g., functional groups, chemical class, derivation, roles).
  \item \textbf{Score 8–9}: Very good — minor omissions or slight rewording, but the core structure-level and functional meaning is intact.
  \item \textbf{Score 6–7}: Reasonable — generally correct but may lack specific details (e.g., derivation or one functional role). Possibly vague phrasing.
  \item \textbf{Score 4–5}: Partial — captures the general category or one function but omits multiple important details or shows misunderstanding in phrasing.
  \item \textbf{Score 2–3}: Poor — vague, generic, or scientifically weak. May refer to the wrong compound type or confuse structural features.
  \item \textbf{Score 1}: Completely incorrect or irrelevant.
\end{itemize}

Only output a \textbf{single line} in the following format:  
\texttt{Score: [1-10]}

\end{tcolorbox}
\end{figure*}

\vspace{0.2em}  % 手动减少段落与 box 之间的空隙
\noindent

\begin{figure*}[!htb]
\begin{tcolorbox}[
  colback=white,
  colframe=black,
  arc=2mm,
  boxrule=0.3pt,
  % breakable,
  % enhanced jigsaw,
  before skip = 1.5mm,  % ← 这是控制“之前”的空白
  width=\textwidth,
]

\textbf{One-pass Self-Consistency: }

Below is a full prompt about the reasoning task, which includes the ICL examples and a new test case. \textbf{Your task is:}
\begin{enumerate}
  \item Read the full prompt to understand the task and identify:  
  \quad 1) the example input-output pairs  
  \quad 2) the specific input question to answer.
  \item Analyze these example pairs and generate a series of rules that explains how each input is transformed to its corresponding output.
  \item Then, apply those rules to the final test question and output the answer.
  \item Return your answer in the following format:
\end{enumerate}

\begin{flushleft}
\texttt{<rules>} \\
\texttt{Rule 1: ...} \\
\texttt{Rule 2: ...} \\
\texttt{Rule 3: ...} \\
\texttt{...} \\
\texttt{</rules>} \\[0.5em]
\texttt{<answer>} \\
\texttt{\{\{your answer\}\}} \\
\texttt{</answer>} \\[0.5em]

\textbf{Full prompt:} \texttt{\{full\_prompt\}}
\end{flushleft}
\end{tcolorbox}
\end{figure*}

\begin{figure*}[!htb]
\begin{tcolorbox}[
% float*=false,        % 禁止浮动（确保在当前位置渲染）
  colback=white,
  colframe=black,
  arc=2mm,
  boxrule=0.3pt,
  % enforce breakable,
  % enhanced jigsaw,
  before skip = 1.5mm,  % ← 这是控制“之前”的空白
  after skip = 1em,
  width=\textwidth,
]

\textbf{Universal Majority Voting with Self-Consistency: }

You are given a reasoning task prompt and multiple candidate responses to the question in that prompt. \textbf{Your task is:}
\begin{enumerate}
  \item Read the full prompt carefully to understand the question being asked.
  \item Examine all the candidate responses and determine whether any of them form a majority consensus.
  \begin{itemize}
    \item A majority exists if \textbf{any single response appears more than any other} (either verbatim or semantically equivalent).
    \item In case of a tie (e.g., all responses differ or two responses appear with equal frequency), consider that no majority exists.
  \end{itemize}
  \item If a majority exists, return that response as the final answer.
  \item If no majority exists, then select the \textbf{most reasonable and task-appropriate} response based on the prompt.
\end{enumerate}

\textbf{Candidate responses:} \texttt{\{responses\}} \\
\textbf{Full prompt:} \texttt{\{full\_prompt\}} \\
\textbf{Return your final answer using \textbf{exactly} the following format:}
\begin{flushleft}
\texttt{majority\_found: [yes or no]} \\
\texttt{selected\_response: \{full response content\}}
\end{flushleft}

\textbf{Example:}
\begin{flushleft}
\texttt{majority\_found: yes} \\
\texttt{selected\_response: This is the most common (or semantically equivalent) response and correctly answers the question.}
\end{flushleft}

\end{tcolorbox}
\end{figure*}

% \clearpage
% \textbf{Hypothesis Refinement}
\begin{figure*}[!htb]
\begin{tcolorbox}[
  colback=white,
  colframe=black,
  arc=2mm,
  boxrule=0.3pt,
  % breakable,
  before skip = 1.5mm,  % ← 这是控制“之前”的空白
  % enforce breakable,
  % enhanced jigsaw,
  width=\textwidth
]
% Prompt 1: Hypothesis Induction
\textbf{Hypothesis Induction Prompt}

Below is a full prompt about the reasoning task, which includes the ICL examples that you should learn from. \textbf{Your task is:}
\begin{enumerate}
  \item Read the full prompt to understand the task and identify the example input-output pairs.
  \item Analyze these example pairs and generate a series of rules that explains how each input is transformed to its corresponding output.
  \item Provide as much detail as possible in the rules, such as elaborating on the specific mapping.\{note\}
  \item Return your rules in the following format (each rule on its own line):
\end{enumerate}
\begin{verbatim}
<hypothesis>
Rule 1: ...
Rule 2: ...
Rule 3: ...
...
</hypothesis>

Full prompt:
{full_prompt}
\end{verbatim}
\end{tcolorbox}
\end{figure*}

% Prompt 2: Hypothesis Application (Chemistry/General)
\begin{figure*}[!htb]
\begin{tcolorbox}[
  colback=white,
  colframe=black,
  arc=2mm,
  boxrule=0.3pt,
  % breakable,
  % enforce breakable,
  % enhanced jigsaw,
  width=\textwidth
]
\textbf{Hypothesis Application Prompt (General)}

\textbf{Task Description:}
{task\_description}

Please apply the given hypothesis to the given list of inputs. Ensure that you provide the actual output for each input. Do not give a program, partial output, or placeholder.

\textbf{Hypothesis:}
{hypothesis}

\textbf{Input:}
{icl\_in}

Format your output as follows:
\begin{verbatim}
<output>
Output 1: ...
Output 2: ...
...
</output>
\end{verbatim}
\end{tcolorbox}
\end{figure*}

% Prompt 3: DNA Table (Single Instance)
\begin{figure*}[!htb]
\begin{tcolorbox}[
  colback=white,
  colframe=black,
  arc=2mm,
  boxrule=0.3pt,
  % breakable,
  % enforce breakable,
  % enhanced jigsaw,
  width=\textwidth
]
\textbf{DNA Table Prompt}

Below is a full prompt about the reasoning task, which includes the question that you should give the corresponding answer. \textbf{Your task is:}
\begin{enumerate}
  \item Read the full prompt to understand the task and identify the specific input question to answer.
  \item Based on your understanding of the given rules, generate the corresponding output for the question.
\end{enumerate}

\textbf{Rules:}
{hypothesis}

Full prompt:
{x}

Enclose your answer with \texttt{<answer></answer>} tags.
\end{tcolorbox}
\end{figure*}

% Prompt 4: DNA Translation/Transformation Code
\begin{figure*}[!htb]
\begin{tcolorbox}[
  colback=white,
  colframe=black,
  arc=2mm,
  boxrule=0.3pt,
  % breakable,
  % enforce breakable,
  after skip=0pt,
  % enhanced jigsaw,
  width=\textwidth
]
\textbf{DNA Translation/Transformation as Python Code Prompt}

Convert the following hypothesis into a Python function called \texttt{apply} that takes a string input and returns the transformed output. 
The function should implement the rules described in the hypothesis. Make sure to handle all the transformations correctly.

\textbf{Task Description:}
{self.task\_description}

\textbf{Hypothesis:}
{hypothesis}

Your function should follow this template:
\begin{verbatim}
def apply(input_str):
    # Implementation based on the hypothesis rules
    # ...
    return result
\end{verbatim}

Return ONLY the Python code without any explanation or markdown formatting.
\end{tcolorbox}
\end{figure*}
\vspace{3em}  % ← 你可以根据需求调成 1em, 0.8em 等
\noindent

% Prompt 5: Hypothesis Refinement
\begin{figure*}[!htb]
\begin{tcolorbox}[
  colback=white,
  colframe=black,
  arc=2mm,
  boxrule=0.3pt,
  % breakable,
  before skip=0pt,  % ← 这是控制“之前”的空白
  % enforce breakable,
  % enhanced jigsaw,
  width=\textwidth,
]
\noindent\textbf{Hypothesis Refinement Prompt}

You are given a candidate hypothesis that attempts to explain how each input is transformed into its output. A hypothesis consists of rules that explain how the inputs are mapped to the outputs. Your goal is to revise this hypothesis so it fully accounts for any discrepancies. You may add new rules, modify existing ones, or remove inaccurate ones. You can also propose a completely new hypothesis.

\textbf{Context:} {self.task\_description}

\textbf{Current Hypothesis:} {hypothesis}

\textbf{Input:} {icl\_in}

\textbf{Model Output:} {generated\_output}

\textbf{Expected Output:} {expected\_output}

\textbf{Steps:}
\begin{enumerate}
  \item List the exact differences between Model Output and Expected Output.
  \item For each difference, identify which existing rule (if any) fails to cover it.
  \item Revise existing rules or introduce new rules to fix these gaps.
  \item Ensure the rules clearly state how the input is mapped into output in a detailed manner.\{note\}
\end{enumerate}

Output only the refined hypothesis—do not solve the original task.

Format your output as follows:
\begin{verbatim}
<new_hypothesis>
Rule 1: ...
Rule 2: ...
Rule 3: ...
...
</new_hypothesis>
\end{verbatim}
\end{tcolorbox}

% Prompt 6: Apply Final Hypothesis
\begin{tcolorbox}[
  colback=white,
  colframe=black,
  arc=2mm,
  boxrule=0.3pt,
  % enhanced jigsaw,
  width=\textwidth,
]
\textbf{Final Hypothesis Application Prompt}

Below is a full prompt about the reasoning task, which includes the question that you should give the corresponding answer. \textbf{Your task is:}
\begin{enumerate}
  \item Read the full prompt to understand the task and identify the specific input question to answer.
  \item Based on your understanding of the given rules, generate the corresponding output for the question.
\end{enumerate}

\textbf{Rules:}
{hypothesis}

Full prompt:
{x}

Enclose your answer with \texttt{<answer></answer>} tags.
\end{tcolorbox}
\end{figure*}

\section{Complete Results on Chemistry Tasks}
We provide the full results on Chemistry Tasks that reports all the metrics in table \ref{tab:chem_claude}, table \ref{tab:chem_gemini}, and table \ref{tab:chem_gpt4.1}.
\begin{table*}[htbp]
  \centering
  
    % \begin{tabularx}{\linewidth}{C{5em} C{6em} C{8.5em} C{8.5em} C{6em} C{6em} C{6em}}
    \resizebox{\linewidth}{!}{
    \begin{tabular}{cccccc}

  \toprule
  \textbf{Task} & \textbf{Metric} &
  \makecell{\textbf{Implicit} \\ \textbf{Inductive} \\ \textbf{Reasoning}} &
  \makecell{\textbf{Explicit} \\ \textbf{Inductive } \\ \textbf{Reasoning}} &
  \makecell{\textbf{Self-} \\ \textbf{Consistency}} &
  \makecell{\textbf{Hypothesis} \\ \textbf{Refinement}} \\
  \midrule

    \multirow{8}[2]{*}{\makecell{Molecule \\ Design}} & exact\_match & 0.17  & \underline{0.23 } & \underline{0.23 } & \textcolor[rgb]{ .216,  .235,  .263}{\textbf{0.27 }} \\
    \multicolumn{1}{c}{} & bleu  & \underline{0.41}  & \textcolor[rgb]{ .216,  .235,  .263}{0.36 } & \textcolor[rgb]{ .216,  .235,  .263}{0.19 } & \textcolor[rgb]{ .216,  .235,  .263}{\textbf{0.71 }} \\
    \multicolumn{1}{c}{} & levenshtein (↓) & \underline{70.87}  & \textcolor[rgb]{ .216,  .235,  .263}{84.70 } & \textcolor[rgb]{ .216,  .235,  .263}{173.47 } & \textcolor[rgb]{ .216,  .235,  .263}{\textbf{26.30 }} \\
    \multicolumn{1}{c}{} & validity & 0.70  & \underline{0.77 } & \textcolor[rgb]{ .216,  .235,  .263}{\textbf{0.80 }} & \textcolor[rgb]{ .216,  .235,  .263}{0.70 } \\
    \multicolumn{1}{c}{} & maccs\_sims & 0.81  & \textcolor[rgb]{ .216,  .235,  .263}{0.75 } & \underline{0.84 } & \textcolor[rgb]{ .216,  .235,  .263}{\textbf{0.89 }} \\
    \multicolumn{1}{c}{} & rdk\_sims & \textbf{0.81 } & \textcolor[rgb]{ .216,  .235,  .263}{0.69 } & \textcolor[rgb]{ .216,  .235,  .263}{0.69 } & \underline{0.76 } \\
    \multicolumn{1}{c}{} & morgan\_sims & 0.62  & \textcolor[rgb]{ .216,  .235,  .263}{0.64 } & \underline{0.66 } & \textcolor[rgb]{ .216,  .235,  .263}{\textbf{0.73 }} \\
    \multicolumn{1}{c}{} & fcd (↓) & \underline{12.82}  & \textcolor[rgb]{ .216,  .235,  .263}{13.87 } & \textcolor[rgb]{ .216,  .235,  .263}{\textbf{12.46 }} & \textcolor[rgb]{ .216,  .235,  .263}{13.22 } \\
    \midrule
    
    \multirow{7}[2]{*}{\makecell{Molecule \\ Caption}} & bleu2 & 0.20  & \textcolor[rgb]{ .216,  .235,  .263}{0.22 } & \textcolor[rgb]{ .216,  .235,  .263}{\textbf{0.39 }} & \underline{0.24}  \\
    \multicolumn{1}{c}{} & bleu4 & \textcolor[rgb]{ .122,  .137,  .161}{0.14 } & \textcolor[rgb]{ .216,  .235,  .263}{0.15 } & \textcolor[rgb]{ .216,  .235,  .263}{\textbf{0.29 }} & \underline{0.17}  \\
    \multicolumn{1}{c}{} & rouge\_1 & \textcolor[rgb]{ .122,  .137,  .161}{0.33 } & \textcolor[rgb]{ .216,  .235,  .263}{0.24 } & \textcolor[rgb]{ .216,  .235,  .263}{\textbf{0.48 }} & \underline{0.40}  \\
    \multicolumn{1}{c}{} & rouge\_2 & \textcolor[rgb]{ .122,  .137,  .161}{0.18 } & \textcolor[rgb]{ .216,  .235,  .263}{0.12 } & \textcolor[rgb]{ .216,  .235,  .263}{\textbf{0.29 }} & \underline{0.23}  \\
    \multicolumn{1}{c}{} & rouge\_l & \textcolor[rgb]{ .122,  .137,  .161}{0.25 } & \textcolor[rgb]{ .216,  .235,  .263}{0.19 } & \textcolor[rgb]{ .216,  .235,  .263}{\textbf{0.38 }} & \underline{0.31}  \\
    \multicolumn{1}{c}{} & meteor\_score & \textcolor[rgb]{ .122,  .137,  .161}{0.39 } & \textcolor[rgb]{ .216,  .235,  .263}{0.23 } & \textcolor[rgb]{ .216,  .235,  .263}{\textbf{0.44 }} & \underline{0.42}  \\
    \multicolumn{1}{c}{} & LLM as judge  & 67.70  & {54.00 } & \underline{69.70 } & \textbf{72.70 } \\
    \midrule
    
    \makecell{Reaction Prediction}
    & accuracy & \textbf{44.44 } & 19.23  & 20.83  & \underline{28.00}  \\
    \midrule
    smiles2formula
      & accuracy & 0.00  & 0.00  & 0.00  & 0.00  \\
    smiles2iupac & accuracy & 0.00  & 0.00  & 0.00  & 0.00  \\
    iupac2smiles & accuracy & \textbf{14.29 } & \underline{4.55}  & 0.00  & 4.17  \\
    iupac2formula & accuracy & 0.00  & \textbf{6.67 } & \underline{3.33}  & \underline{3.33}  \\
    \bottomrule
    % \end{tabularx}%
    \end{tabular}
    }
  \caption{Performance of the \textbf{Claude-3.5-Haiku} on Chemistry Tasks}
  \label{tab:chem_claude}%
\end{table*}%

\begin{table*}[htbp]
  \centering
  
    % \begin{tabularx}{\linewidth}{C{5em} C{6em} C{8.5em} C{8.5em} C{6em} C{6em} C{6em}}
    \resizebox{\linewidth}{!}{
    \begin{tabular}{cccccc}
  \toprule
  \textbf{Task} & \textbf{Metric} &
  \makecell{\textbf{Implicit} \\ \textbf{Inductive} \\ \textbf{Reasoning}} &
  \makecell{\textbf{Explicit} \\ \textbf{Inductive } \\ \textbf{Reasoning}} &
  \makecell{\textbf{Self-} \\ \textbf{Consistency}} &
  \makecell{\textbf{Hypothesis} \\ \textbf{Refinement}} \\
  \midrule

    \multirow{8}[2]{*}{\makecell{Molecule \\ Design}} & exact\_match & \textbf{0.30}  & 0.20 & 0.20 & \underline{0.23} \\
    \multicolumn{1}{c}{} & bleu  & \textbf{0.75} & 	\underline{0.71} & 	0.70 & 	\textbf{0.75} \\
    \multicolumn{1}{c}{} & levenshtein (↓) & \underline{25.37} &	27.93 &	26.37 &	\textbf{24.03}  \\
    \multicolumn{1}{c}{} & validity & 0.87 &	\textbf{1.00} 	& \underline{0.93} 	& \underline{0.93}  \\
    \multicolumn{1}{c}{} & maccs\_sims & \textbf{0.92} &	0.87 	& \underline{0.91} 	& 0.87  \\
    \multicolumn{1}{c}{} & rdk\_sims & \underline{0.80} &	0.74 &	\textbf{0.82} &	0.78  \\
    \multicolumn{1}{c}{} & morgan\_sims & \textbf{0.75} &	0.69 &	\underline{0.72} &	0.67  \\
    \multicolumn{1}{c}{} & fcd (↓)&8.16 &	\textbf{7.08} &	7.97 	& \underline{7.43}    \\
    \midrule
    
    \multirow{7}[2]{*}{\makecell{Molecule \\ Caption}} & bleu2 & \underline{0.42} &	\textbf{0.49} 	&\textbf{ 0.49} &	0.20  \\
    \multicolumn{1}{c}{} & bleu4 & 0.32 	&\underline{0.38} 	&\textbf{0.39} &	0.15 \\
    \multicolumn{1}{c}{} & rouge\_1 & \underline{0.55}& 	\underline{0.55} 	&\textbf{0.57} &	0.38  \\
    \multicolumn{1}{c}{} & rouge\_2 & 0.36 &	\underline{0.38} &	\textbf{0.39} &	0.24  \\
    \multicolumn{1}{c}{} & rouge\_l & 0.44 &	\underline{0.46} &	\textbf{0.48} &	0.31  \\
    \multicolumn{1}{c}{} & meteor\_score & \textbf{0.57} 	&0.52 	&\underline{0.54} &	0.48  \\
    \multicolumn{1}{c}{} & LLM as judge  & \textbf{66.30} &	59.00 &	\underline{65.70} &	\textbf{66.30} \\
    \midrule
    
    \makecell{Reaction Prediction}
    & accuracy & 22.22 &	17.86 &	\underline{25.00} &	\textbf{32.14} \\
    \midrule
    smiles2formula
      & accuracy & \textbf{13.33} &	6.67 &	\underline{10.00} &	\underline{10.00} \\
    smiles2iupac & accuracy & 0.00 &	0.00 &	0.00 	&0.00  \\
    iupac2smiles & accuracy & \textbf{17.39} &	4.35 &	5.00 	&\underline{13.04}  \\
    iupac2formula & accuracy & \textbf{23.33}&	\underline{13.33}&	\textbf{23.33}	&\textbf{23.33} \\
    \bottomrule
    % \end{tabularx}%
    \end{tabular}
    }
  \caption{Performance of the \textbf{GPT-4.1} on Chemistry Tasks}
  \label{tab:chem_gpt4.1}%
\end{table*}%

\begin{table*}[htbp]
  \centering
  
    % \begin{tabularx}{\linewidth}{C{5em} C{6em} C{8.5em} C{8.5em} C{6em} C{6em} C{6em}}
    \resizebox{\linewidth}{!}{
    \begin{tabular}{cccccc}

  \toprule
  \textbf{Task} & \textbf{Metric} &
  \makecell{\textbf{Implicit} \\ \textbf{Inductive} \\ \textbf{Reasoning}} &
  \makecell{\textbf{Explicit} \\ \textbf{Inductive } \\ \textbf{Reasoning}} &
  \makecell{\textbf{Self-} \\ \textbf{Consistency}} &
  \makecell{\textbf{Hypothesis} \\ \textbf{Refinement}} \\
  \midrule

    \multirow{8}[2]{*}{\makecell{Molecule \\ Design}} & exact\_match & \textbf{0.33} &	\underline{0.27} &	\underline{0.27} &	0.20 \\
    \multicolumn{1}{c}{} & bleu  & 0.73 &	\textbf{0.79}& 	\textbf{0.79} &	\underline{0.76} \\
    \multicolumn{1}{c}{} & levenshtein (↓) & 27.90 	&\underline{25.27} &	\textbf{22.50 }	&26.67  \\
    \multicolumn{1}{c}{} & validity &  \underline{0.80} 	&0.77 &	\textbf{0.90} 	&0.73  \\
    \multicolumn{1}{c}{} & maccs\_sims & \textbf{0.95} 	&\underline{0.94} &	\underline{0.94} &	0.81 \\
    \multicolumn{1}{c}{} & rdk\_sims & \textbf{0.89} 	&0.86 &	\underline{0.87} 	&0.82 \\
    \multicolumn{1}{c}{} & morgan\_sims & \textbf{0.85} 	&0.77 	&\underline{0.80} &	0.72 \\
    \multicolumn{1}{c}{} & fcd (↓)& \underline{8.19} 	&8.89 &	\textbf{6.26} 	&10.56 \\
    \midrule
    
    \multirow{7}[2]{*}{\makecell{Molecule \\ Caption}} & bleu2 & 0.49 	&\textbf{0.54 }&	\underline{0.51} 	&0.42 \\
    \multicolumn{1}{c}{} & bleu4 & 0.38 	&\textbf{0.43} 	&\underline{0.41} &	0.33 \\
    \multicolumn{1}{c}{} & rouge\_1 &  \underline{0.57} 	&\textbf{0.61} &	\textbf{0.61 }	&0.52 \\
    \multicolumn{1}{c}{} & rouge\_2 & 0.38 	&\textbf{0.42} 	&\underline{0.41} &	0.35 \\
    \multicolumn{1}{c}{} & rouge\_l &  0.47 	&\textbf{0.50 }	&\underline{0.49} &	0.43 \\
    \multicolumn{1}{c}{} & meteor\_score & \underline{0.55} &	\textbf{0.59 }	&\textbf{0.59} &	0.52 \\
    \multicolumn{1}{c}{} & LLM as judge  & 63.30 	&\underline{67.70} &	\textbf{70.00 }	&65.70 \\
    \midrule
    
    \makecell{Reaction Prediction} & accuracy & \textbf{54.17} &	34.78 &	\underline{39.29} &	32.14 \\
    \midrule
    smiles2formula & accuracy & \textbf{30.00} 	&\underline{20.00} &	\textbf{30.00} &	16.67 \\
    smiles2iupac & accuracy & 0.00 &	0.00 	&\textbf{3.33} &	0.00 \\
    iupac2smiles & accuracy &  20.00 &	40.00 	&\textbf{53.85 }	&\underline{52.94} \\
    iupac2formula & accuracy &  \underline{70.00} 	&60.00 	&\textbf{73.33} &	66.67 \\
    \bottomrule
    % \end{tabularx}%
    \end{tabular}
    }
  \caption{Performance of the \textbf{Gemini-2.5-Flash} on Chemistry Tasks}
  \label{tab:chem_gemini}%
\end{table*}%

\end{document}